# Continuous Authentication Using Mouse Movements, Machine Learning, and Minecraft

Nyle Siddiqui
*Department of Computer Science*
*University of Wisconsin - Eau Claire*
Eau Claire, US
siddiqun8701@uwec.edu

Rushit Dave
*Department of Computer Science*
*University of Wisconsin – Eau Claire*
Eau Claire, US
daver@uwec.edu

Naeem Seliya
*Department of Computer Science*
*University of Wisconsin – Eau Claire*
Eau Claire, US
seliyana@uwec.edu

*Abstract*—Mouse dynamics has grown in popularity as a novel, irreproducible behavioral biometric. Datasets which contain general, unrestricted mouse movements from users are sparse in the current literature. The Balabit mouse dynamics dataset, produced in 2016, was made for a data science competition and despite some of its shortcomings, is considered to be the first publicly available mouse dynamics dataset. Collecting mouse movements in a dull, administrative manner, as Balabit does, may unintentionally homogenize data and is also not representative of real-world application scenarios. This paper presents a novel mouse dynamics dataset that has been collected while 10 users play the video game *Minecraft* on a desktop computer. Binary Random Forest (RF) classifiers are created for each user to detect differences between a specific user's movements and an imposter's movements. Two evaluation scenarios are proposed to evaluate the performance of these classifiers; one scenario outperformed previous works in all evaluation metrics, reaching average accuracy rates of 92%, while the other scenario successfully reported reduced instances of false authentications of imposters.

## I. INTRODUCTION

Social dependency on technology to assist in day-to-day tasks has grown increasingly widespread. E-commerce, online banking, and the use of personal cloud storage have become a part of essential daily routines. User authentication methods have been implemented to ensure that only trusted users on networks and other web-based environments are able to securely store and transmit relevant data. Physiological-based biometrics, such as fingerprint, iris, facial, and voice, have been previously proposed as reliable, irreproducible biometrics for distinct user authentication [1-3]. Unfortunately, these methods lack the ability to continuously authenticate a user after the initial verification stage. After a nefarious user gains the credentials to access an online account or network, there are no additional security layers in place to prevent further security breaches. Therefore, engineering methods to mitigate and respond to these attacks have become a critical issue in cybersecurity.

A user's mouse usage, formally known as mouse dynamics, has recently been proposed as a non-intrusive, effective behavioral biometric for user authentication [4-7]. Mouse dynamics exploits the subtle yet distinct differences in fine motor behavior between humans when using a mouse to uniquely identify between users. Moreover, it requires less sensitive data from the user as opposed to other behavioral biometrics [8]. Mouse dynamics is the easiest and most abundant modality through which data can be collected in the context of the previously discussed web-based environments, thus making it a prime candidate for future online behavioral-based user authentication schemes as seen in [9, 10]. In addition to mouse dynamics' recent exposure, machine and deep learning algorithms have had a dominant presence in large data-driven environments and, in many cases, outperform conventional statistical methods in a variety of these problem spaces [11-14].

In this paper, we train a machine learning algorithm, namely a Random Forest (RF) algorithm, in a binary classification setting on a novel mouse dynamics dataset for user authentication. This dataset was collected from 10 participants as they played the video game *Minecraft* on a desktop for 20 minutes. Each participant was not given a task and thus were free to engage in any activity of their choosing for the session. We hypothesize that collecting mouse dynamics in this unique, high-intensity environment - as opposed to the slow and restrictive administrative tasks that are usually given to participants during mouse dynamics data collection – will improve the quality of user classifiers and present novel problems to further solve and improve the field. The novel contributions of this paper are as follows:

- Introduce a novel mouse dynamics dataset containing data from 10 users while they played the video game *Minecraft*. Dataset is available at: https://github.com/NyleSiddiqui/Minecraft-Mouse-Dynamics-Dataset.

- Develop multiple Random Forest (RF) classifiers for each user and evaluate it using two different scenarios in order to determine the optimal user authentication scheme.

- Data availability and reproducibility were emphasized in the process of this experiment to

foster more future work; a lacking aspect in current mouse dynamics works.

## II. BACKGROUND AND RELATED WORK

Behavioral biometrics in general has been receiving increased focus in cybersecurity related research. Additionally, many machine and deep learning algorithms have been applied to this domain due to their superior ability to efficiently identify implicit patterns within large amounts of data, extract important features from raw data, and leveraging the spatiotemporal nature of behavioral data. However, it is important to note the variance in mouse dynamics data collection methods and its impact on the dissemination of results. Currently, there are two common types of mouse dynamics data in the literature: specific data and general data. Specific data is defined as data collected from participants engaging in an assigned task on a desktop/web application [15, 16], whereas general data is generated continuously while users interact with and perform unspecified daily tasks on their computer [17, 18]. Our data is a combination of the two as the data is collected using *Minecraft,* however, the user is free to do whatever they choose and thus generates general data.

Furthermore, raw mouse dynamics data is not sufficiently granular for prediction models to accurately base their authentication decisions on it. Therefore, detailed features must be extracted from the raw data and appended to the dataset during model training. Due to the relative infancy of the field, there are not yet any concretely established extractable features that have been proven to generally outperform all others. Our feature extraction process is further discussed in Section III.

[16] was one of the very first examples of mouse dynamics for user authentication and assisted in laying fundamental guidelines for future experiments. In order to verify the validity of mouse dynamics as a viable behavioral biometric, the authors compare the universality, uniqueness, permeance, collectability, performance, acceptability, and possibility of circumvention to other behavioral modalities. Further, they propose the partitioning of data by using "strokes" – defined as the set of points between two mouse clicks, in addition to various suggested spatial and temporal feature extractions. [16] set the stage for future mouse dynamics research by introducing a novel behavioral biometric, presenting an optimal feature selection process, providing concrete examples of the benefits of mouse dynamics over other conventional behavioral biometrics, and finally proving their results using non-parametric mathematical estimation methods; these methods reaching equal error rates as low as 0.2%. Recent research has expanded on this paper by utilizing machine learning to further optimize user classification and intrusion detection accuracies.

In [19], the authors tested various standalone recurrent neural networks (RNN) and convolutional neural networks (CNN) on 2 public mouse dynamics datasets for user authentication: the TWOS dataset and the Balabit Mouse Dynamics Challenge dataset. The purpose was to compare the efficacy of deep learning algorithms for user authentication using mouse dynamics against a standard, baseline machine learning classification algorithm: the support vector machine (SVM). The authors hypothesized that the more complex RNN and CNN models would outperform the SVM in user mouse dynamics classification due to their aforementioned ability in utilizing temporal and spatial features of the data. Their results and conclusions supported their hypothesis; however, they note that many limitations in this article occur from the public datasets used rather than the models themselves. One of the most glaring problems is that the public datasets are collected by users that partake in specific, identical tasks. This raises the question whether these models are truly learning long-term user behavior or simply the relationship between each user and this single specific task. Another limitation is the simplicity that arises in the data from these dull tasks may not represent the entropy in real-world implementations. This further reiterates the need for newer, unorthodox datasets.

The authors of [20] use the browser-based game *Perfect Piano* for their data collection. They extract 87 separate features from the raw mouse data for both continuous user authentication and anomaly detection. Before extraction, they separate their data based on the mouse action taken: mouse movements, point and click, and drag and drop. The authors also reference [21] in which, similarly, a deep neural network was developed for user authentication from mouse dynamics collected from the game *League of Legends*. This paper did note that mouse dynamics data collected in video game environments are susceptible to variability depending on what stage in the game the data was collected, thereby impacting model accuracy and adaptability. [20] goes on to propose using a CNN for anomaly detection and user authentication and found that this model retained the highest precision, recall, and accuracy when compared to other baseline machine learning algorithms. While both [20] and [21] do collect data in the previously mentioned high-intensity game environments, many of these datasets are not published for public use - hence our strong intent to publish publicly available data.

## III. METHEODOLOGIES

### A. Dataset Description

This dataset collected 10 users mouse dynamics while they played the video game *Minecraft* on a desktop computer for 20 minutes. The version of *Minecraft* used was the free trial version available on all Windows desktop operating systems. Each user played on the exact same desktop which ran a 64-bit Windows 10 operating system with an Intel i7-8700 CPU clocked at 3.20GHz, 16GB of DDR4 RAM, and NVIDIA GeForce GT 730 graphics card connected to a 1080p Dell P2217H monitor. Each user also used the same Dell MS116 USB Optical Mouse, Dell KB216 Wired Keyboard, and were subjected to default *Minecraft* settings and key binds during each session.

A Python program ran in the background for the allotted 20 minutes and collected data from each user during their session. This program is also available in the GitHub repository. The raw data recorded from each user was written and saved to a text file, where each line represented a unique mouse event: either a mouse movement, click, or scroll. Each line of text contained 5 fields: (*Timestamp, X, Y, Event Type, ID*) where *Timestamp* is the UNIX timestamp during an event, X and Y are the x-coordinate and y-coordinate, respectively, of the cursor during an event, *Event Type* is the type of mouse event (movement, click, or scroll), and *ID* is the corresponding user ID the data was collected from. A visual representation of a user's session can be observed in Fig. 1.

*B. Feature Extraction and User Profiles*

In order to clean the data, duplicate entries were omitted. Following [16] and [22], we define a mouse action as a sequence of a certain number of consecutive mouse events. In this paper, we choose this number to be 10 to ensure there are sufficient data remaining to properly train the RF. That is to say, a mouse action is an aggregation of 10 separate, consecutive mouse events (Fig. 2). These mouse actions are the data that additional features are extracted from. We follow the feature extraction process as described in [22] for each user's dataset individually. These features include, but are not limited to, the aggregate functions of a mouse action's velocity, acceleration, and movement direction. After the individual feature extraction of each user's data, the data are split into training and testing sets with a train-test split of 70-30. Finally, each user's data is concatenated into a master training and testing dataset.

Previous research has suggested that two-class (binary) classifiers outperform one-class classifiers when trained on behavioral biometric-based data [23]. While it is difficult to procure negative examples in this regard with most behavioral biometrics, such as keystroke dynamics [24], it is relatively easy for mouse dynamics problems to be framed in this manner. Therefore, for each user, we train a separate RF classifier to determine whether inputted mouse actions belong to that specific user or not. Training a separate RF classifier for each user ensures that the classifier is more adept at anomaly and intruder detection. We use the Scikit-learn Python library (version 0.24.2) to create and evaluate the RFs.

Training and testing sets contain two classes: the genuine user, which is the positive class, and the imposter user, which is the negative class. Each user-specific dataset is perfectly balanced - they contain an equal number of genuine and imposter instances to mitigate bias within the RF classifier. For example, if user 0 possessed *n* number of mouse actions, their specific dataset would contain *n* number of imposter mouse actions selected equally from the rest of the nine users. In other words, each user-specific dataset to train each user-specific classifier would have *N* examples: *n* genuine examples and *n* imposter examples – these imposter examples being comprised of *n*/9 mouse actions selected from each imposter user.

Fig. 2. Mouse action creation process. Removed duplicates highlighted in red.

IV. SYSTEM EVALUATION

We propose two separate evaluation scenarios as in [22]. In Scenario A, the training dataset is exclusively used to train and test each binary RF classifier. In Scenario B, the classifier is trained on the training dataset and tested on the testing dataset. Scenarios A and B measure each classifier's performance with respect to in-distribution data and out-of-distribution, respectively.

To measure the performance of each individual RF, we calculate the following evaluation metrics: accuracy, false-negative rate (FNR), false positive rate (FPR), and equal error rate (EER). Accuracy is the simplest evaluation metric and simply represents the percentage of correctly classified data samples. While false positives (authenticating an imposter as a trusted user) are much more detrimental to overall system security than false negatives (not authenticating a trusted user) [25], we still aim to optimally reduce both instances of false predictions; note that false positives should still be preferably mitigated. FNR is also known as the false rejection rate, or the proportion of genuine user instances that were falsely classified as an imposter. Contrastingly, FPR, also knows as false acceptance rate, is the porportion of imposter samples that were incorrectly classified as a trusted user. EER, or equal error rate, expressses the error rate for a given classifier's threshold in which the FPR and FNR are equal

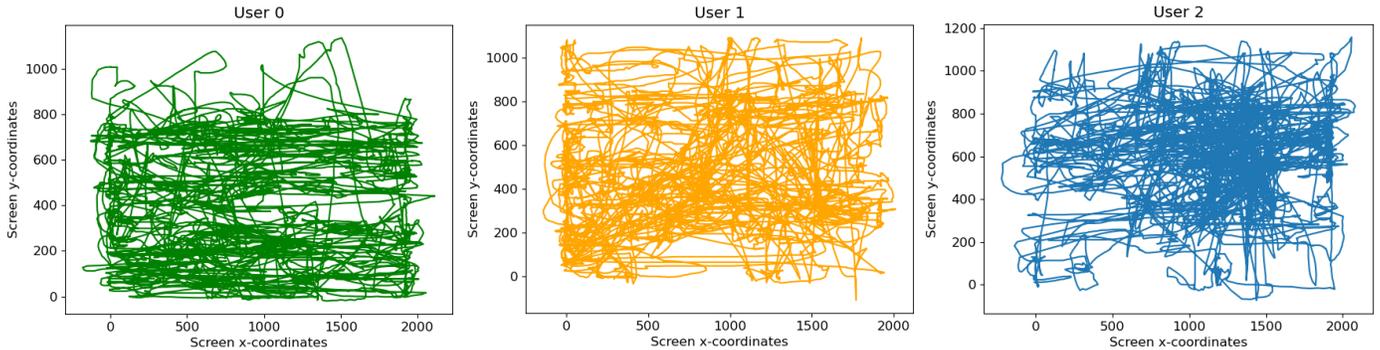

Fig. 1. Mouse movement maps from three users' gaming sessions.

TABLE I. Results from Scenario A for each user's binary RF classifier. Highest and lowest values are highlighted in green and red, respectively.

| User | # of Genuine Actions | ACC | FNR (Threshold = 0.5) | FPR (Threshold = 0.5) | EER (User-Specific Threshold) |
|---|---|---|---|---|---|
| 0 | 32953 | 0.9263 | 0 | 0.1473 | 0.002063 |
| 1 | 39684 | 0.9327 | 0 | 0.1186 | 0.001890 |
| 2 | 36685 | 0.9373 | 0 | 0.1608 | 0.002126 |
| 3 | 51510 | 0.9236 | 0 | 0.1528 | 0.002025 |
| 4 | 39395 | 0.9034 | 0 | 0.1932 | 0.002132 |
| 5 | 32931 | 0.9339 | 0 | 0.1321 | 0.001937 |
| 6 | 28568 | 0.9305 | 0 | 0.1390 | 0.002380 |
| 7 | 42903 | 0.9311 | 0 | 0.1378 | 0.001118 |
| 8 | 31169 | 0.9239 | 0 | 0.1522 | 0.001925 |
| 9 | 41186 | 0.9300 | 0 | 0.1401 | 0.0007283 |
| Avg. | | 0.9273 | 0 | 0.1439 | 0.001832 |
| Std. | | 0.0094 | 0 | 0.0205 | 0.0005075 |

### A. Scenario A Results

In this subsection, we describe the results observed when training and testing each RF classifier exclusively with the master training dataset, as seen in Table I. Cumulatively, the master training dataset contains 376,984 distinct mouse actions.

Overall, we observed higher overall average accuracy across all classes when compared to results from [22] with an average accuracy rate of 92.73%. Moreover, lower average FNR, FPR, and EER values were observed at 0%, 14.39%, and 0.1832%, respectively. Note that EER values were obtained with user-specific thresholds. This means that optimal threshold values were found individually for each RF classifier before calculating EER. Furthermore, the standard deviations of these values were also lower than observed in [22]. In detail, User 4 was observed to have the lowest accuracy at 90.34% as well as the highest FPR with nearly 20% of imposter actions being incorrectly authenticated. User 2 had the highest accuracy rate at 93.73% and User 1 had the lowest FPR at 11.86%. User 9 observed the lowest EER with a value of 0.07% and even the highest EER value from User 6 was quite negligible at 0.23%. Since the feature extraction process and classifier construction were near identical to [22], this could be an indication that our dataset, collected in a free, high-intensity environment, can result in higher quality and more descriptive data.

### B. Scenario B Results

In this subsection, we describe the results observed when training each RF classifier with the training dataset and testing on the testing dataset, as seen in Table II. Cumulatively, the test dataset contains 92,899 distinct mouse actions.

TABLE II. Results from Scenario B for each user's binary RF classifier.

| User | # of Genuine Actions | ACC | FNR (Threshold = 0.5) | FPR (Threshold = 0.5) | EER (User-Specific Threshold) |
|---|---|---|---|---|---|
| 0 | 8334 | 0.6550 | 0.6489 | 0.1702 | 0.3649 |
| 1 | 9474 | 0.6153 | 0.6222 | 0.2270 | 0.3939 |
| 2 | 9207 | 0.5670 | 0.7457 | 0.1862 | 0.4338 |
| 3 | 12512 | 0.6229 | 0.6107 | 0.2347 | 0.3914 |
| 4 | 9784 | 0.6827 | 0.4653 | 0.2873 | 0.3447 |
| 5 | 7920 | 0.5780 | 0.7704 | 0.1627 | 0.4219 |
| 6 | 6903 | 0.6719 | 0.5593 | 0.2172 | 0.3544 |
| 7 | 10746 | 0.6139 | 0.6388 | 0.2242 | 0.4125 |
| 8 | 7562 | 0.5790 | 0.7113 | 0.1871 | 0.4228 |
| 9 | 10182 | 0.5740 | 0.7078 | 0.2132 | 0.4272 |
| Avg. | | 0.6160 | 0.6480 | 0.2110 | 0.3967 |
| Std. | | 0.0423 | 0.0915 | 0.0366 | 0.0323 |

Our results in Scenario B were more moderate compared to Scenario A. They are also not as comparable to [22] as we only used mouse movement events to create actions while [22] found using only mouse click events/actions to yield the best results. However, we did obtain a lower FPR by approximately 19% when only using mouse movements events/actions, which is significant in the context of user authentication. Our results did yield reduced accuracy and increased FNR rates, so optimized trade-offs between these three values should be further explored, especially since the significance of these values vary depending on situational context. There doesn't appear to be any correlation between best and worst performing user classifiers when comparing results from Scenario A, which follows since our interclass deviations are small in both evaluation scenarios.

## V. DISCUSSION AND ANALYSIS

[16] successfully demonstrated the use of mouse dynamics for user authentication by achieving EER values as low as 0.2%. It is important to note, however, that this EER value was only obtainable when a mouse action sequence length of 200 mouse events was used. When using a mouse action sequence length of 10, they observed an EER of 11.8%. Compared to our average EER of 0.2% in Scenario A at a mouse action sequence length of 10, this may again indicate that collecting data in a volatile and high-intensity environment encodes higher quality information in smaller quantities. Furthermore, smaller mouse action sequence lengths may require less computational resources in the context of a future real-world implementation. Our Scenario B EER more closely resembled the EER values when one or two mouse events were used for mouse action creation in [16]. This may be due to the classifiers' difficulty processing never-before-seen data rather than the mouse action sequence length, as seen from the successful results in Scenario A.

Our experiment followed many processes seen in [22] so results are more easily comparable and significant. Firstly, in Scenario A, we reported an average FPR of 14.39% and accuracy of 92.73% – approximately a 10.5% decrease in relation to their FPR of 24.88% and 12.5% increase in relation to their 80.17% accuracy rate. Furthermore, the interclass standard deviations of these two values were significantly less variant in our experiment. These values were achieved in [22] by first finding the most informative mouse event type before creating mouse actions to train the classifiers with. Further improved results may be observed in both Scenario A and B if these same steps are followed in future works with this dataset. Our Scenario B results reported lower FPR at the cost of lower accuracy and higher FNR when compared to Scenario B in [22]. However, interclass accuracy and FPR values here highly variant in [22], whereas our results were smoother across all classes. Despite lower performance, our classifiers would operate in a more consistent manner in a simulated real-world environment, leaving room for future improved results.

## VI. LIMITATIONS AND FUTURE WORK

In practice, a user authentication scheme would want to prioritize minimizing FPR over FNR as false positives (authenticating an imposter as a trusted user) are much more detrimental to overall system security than false negatives (not authenticating a trusted user. Thresholds can be further optimized to achieve this goal. Besides the promising results shown in Scenario A, Scenario B's results indicate that there may be other algorithms that can correctly evaluate out-of-distribution more effectively. While low FPR rates were observed, lower accuracy and higher FNR may highlight the RF classifiers' difficulty to effectively operate on never-before-seen data. As noted in Section II, deep learning algorithms such as long short-term memory recurrent neural networks and convolutional neural networks may be used in the future on this dataset to better fit to testing data. Additionally, previous literature exhibited a significant difference in results when different mouse action sequence lengths and a mixture of all mouse events (point and clicks, drag and drops, not just mouse movements) to form mouse actions were used. Further optimization of these hyperparameters in this paper could lead to improved results.

## VII. CONCLUSION

In this paper, we proposed a novel mouse dynamics dataset collected in a high-intensity environment. This data was specifically collected with the intent of user authentication using mouse dynamics as well as contributing to the limited number of publicly available mouse dynamics datasets. We created binary RF classifiers for each user and tested them against other user's data posing as "imposters" trying to mimic a trusted user's mouse movements. These classifiers were tested using two different evaluation scenarios. Measuring each classifier's ability to perform on in- and out-of-distribution data. While the classifiers did struggle somewhat on out-of-distribution data, low FPR rates were observed in both scenarios as compared to previous literature (0.1439 and 0.2110, respectively), a significant aspect in user authentication schemes.

Overall, our findings support the continued research of mouse dynamics as a viable behavioral biometric for user authentication. If utilized in an intrusion detection system in conjunction with additional alternative behavioral biometrics, accuracy rates may increase above a sufficient threshold to begin implementing and evaluating in real-world scenarios. We will actively continue to collect data from additional users to add to this dataset. This will allow for deeper, more complex algorithms to train on the dataset as well as allow for extra data pre-processing, such as creating mouse actions under different number/types of mouse events.

## ACKNOWLEDGEMNTS

Funding for this project was provided by the University of Wisconsin - Eau Claire's Office of Research and Sponsored Programs. Thanks to the Blugold Supercomputing Cluster (BGSC) for providing computational resources.

## REFERENCES

[1] Siddiqui N., Pryor L., Dave R. (2021) User Authentication Schemes Using Machine Learning Methods—A Review. In: Kumar S., Purohit S.D., Hiranwal S., Prasad M. (eds) Proceedings of International Conference on Communication and Computational Technologies. Algorithms for Intelligent Systems. Springer, Singapore. https://doi.org/10.1007/978-981-16-3246-4_54.

[2] J. Shelton et al., "Palm Print Authentication on a Cloud Platform," 2018 International Conference on Advances in Big Data, Computing and Data Communication Systems (icABCD), 2018, pp. 1-6, doi: 10.1109/ICABCD.2018.8465479.

[3] Akhtar, Z., & Buriro, A. (2021). Multitrait Selfie: Low-Cost Multimodal Smartphone User Authentication. In *Biometric Identification Technologies Based on Modern Data Mining Methods* (pp. 159-175). Springer, Cham.

[4] Shen, C., Cai, Z., Guan, X., Du, Y., & Maxion, R. A. (2013). User Authentication Through Mouse Dynamics. IEEE Transactions on Information Forensics and Security, 8(1), 16-30. doi:10.1109/tifs.2012.2223677.

[5] Antal, M., & Fejer, N. (2020). Mouse dynamics based user recognition using deep learning. Acta Universitatis Sapientiae, Informatica, 12(1), 39-50.

[6] Almalki, S., Chatterjee, P., & Roy, K. (2019). Continuous authentication using mouse clickstream data analysis. In *International Conference on Security, Privacy and Anonymity in Computation, Communication and Storage* (pp. 76-85). Springer, Cham.

[7] A. K., Tiwari, P., & Hossain, M. S. (2020). Predicting users' behavior using mouse movement information: an information foraging theory perspective. *Neural Computing and Applications*, 1-14.

[8] Mason, J., Dave, R., Chatterjee, P., Graham-Allen, I., Esterline, A., & Roy, K. (2020). An Investigation of Biometric Authentication in the Healthcare Environment. Array, 8, 100042. doi:10.1016/j.array.2020.100042

[9] Krátky, Peter & Chuda, Daniela. (2018). Recognition of web users with the aid of biometric user model. Journal of Intelligent Information Systems. 51. 10.1007/s10844-018-0500-0.

[10] Ahmed, A. A. E., & Traore, I. (2007). A new biometric technology based on mouse dynamics. *IEEE Transactions on dependable and secure computing*, 4(3), 165-179.

[11] Strecker S., Van Haaften W., Dave R. (2021) An Analysis of IoT Cyber Security Driven by Machine Learning. In: Kumar S., Purohit S.D., Hiranwal S., Prasad M. (eds) Proceedings of International Conference on Communication and Computational Technologies. Algorithms for Intelligent Systems. Springer, Singapore. https://doi.org/10.1007/978-981-16-3246-4_55.

[12] Ackerson JM, Dave R, Seliya N. Applications of Recurrent Neural Network for Biometric Authentication & Anomaly Detection. Information. 2021; 12(7):272. https://doi.org/10.3390/info12070272

[13] Gunn, Dylan J. et al. "Touch-Based Active Cloud Authentication Using Traditional Machine Learning and LSTM on a Distributed Tensorflow Framework." Int. J. Comput. Intell. Appl. 18 (2019): 1950022:1-1950022:16

[14] Wei, A., Zhao, Y., & Cai, Z. (2019). A deep learning approach to web bot detection using mouse behavioral biometrics. In *Chinese Conference on Biometric Recognition* (pp. 388-395). Springer, Cham.Systems.

[15] Shen, C., Cai, Z., Guan, X., & Maxion, R. (2014). Performance evaluation of anomaly-detection algorithms for mouse dynamics. *computers & security*, *45*, 156-171.

[16] Gamboa, H. & Fred, A. (2004). A behavioral biometric system based on humancomputer interaction. Proc SPIE. 5404. 381-392. 10.1117/12.542625.

[17] Feher, C., Elovici, Y., Moskovitch, R., Rokach, L., & Schclar, A. (2012). User identity verification via mouse dynamics. Information Sciences, 201, 19-36.

[18] Shen, C., Cai, Z., & Guan, X. (2012). Continuous authentication for mouse dynamics: A pattern-growth approach. In *IEEE/IFIP International Conference on Dependable Systems and Networks (DSN 2012)* (pp. 1-12). IEEE.

[19] Chong, P., Elovici, Y., & Binder, A. (2020). User Authentication Based on Mouse Dynamics Using Deep Neural Networks: A Comprehensive Study. IEEE Transactions on Information Forensics and Security, 15, 1086-1101. doi:10.1109/tifs.2019.2930429.

[20] Almalki, S.; Assery, N.; Roy, K. An Empirical Evaluation of Online Continuous Authentication and Anomaly Detection Using Mouse Clickstream Data Analysis. Appl. Sci. 2021, 11, 6083. https://doi.org/10.3390/app11136083.

[21] Wei, A., Zhao, Y., & Cai, Z. (2019). A deep learning approach to web bot detection using mouse behavioral biometrics. In *Chinese Conference on Biometric Recognition* (pp. 388-395). Springer, Cham.Systems.

[22] Antal, M. & Egyed-Zsigmond, E. (2019). Intrusion Detection Using Mouse Dynamics. IET Biometrics. 10.1049/iet-bmt.2018.5126.

[23] Antal, M., & Szabó, L.Z. (2015). An Evaluation of One-Class and Two-Class Classification Algorithms for Keystroke Dynamics Authentication on Mobile Devices. 2015 20th International Conference on Control Systems and Computer Science, 343-350.

[24] Pepa, L., Sabatelli, A., Ciabattoni, L., Monteriù, A., Lamberti, F., & Morra, L. (2020). Stress detection in computer users from keyboard and mouse dynamics. *IEEE Transactions on Consumer Electronics*, *67*(1), 12-19.

[25] N. Zheng, A. Paloski, and H. M. Wang. (2011) "An efficient user verification system via mouse movements," in Proc. ACM Conf. Computer and Communications Security, Chicago, IL, 2011, pp. 139–150.